\documentclass[11pt,a4paper]{article}
\pdfoutput=1
\usepackage{authblk}
\usepackage[hyphens]{url}
\usepackage[hyperref]{emnlp2018}
\usepackage{times}
\usepackage{latexsym}
\usepackage{booktabs}

\aclfinalcopy 

\newcommand\Tstrut{\rule{0pt}{2.3ex}}

\title{Challenges for Toxic Comment Classification:\\An In-Depth Error Analysis}

\author[1]{Betty van Aken}
\author[2]{Julian Risch}
\author[2]{Ralf Krestel}
\author[1]{Alexander L{\"o}ser}
\affil[1]{Beuth University of Applied Sciences Berlin, Germany \protect\\{{\tt \{bvanaken, aloeser\}@beuth-hochschule.de}}}
\affil[2]{Hasso Plattner Institute, University of Potsdam, Germany \protect\\{\tt firstname.lastname@hpi.de}}

\date{}

\begin{document}
\maketitle
\begin{abstract}
Toxic comment classification has become an active research field with many recently proposed approaches.
However, while these approaches address some of the task's challenges others still remain unsolved and directions for further research are needed.
To this end, we compare different deep learning and shallow approaches on a new, large comment dataset and propose an ensemble that outperforms all individual models.
Further, we validate our findings on a second dataset.
The results of the ensemble enable us to perform an extensive error analysis, which reveals open challenges for state-of-the-art methods and directions towards pending future research. 
These challenges include missing paradigmatic context and inconsistent dataset labels.
\end{abstract}

\section{Introduction}

Keeping online conversations constructive and inclusive is a crucial task for platform providers. 
Automatic classification of toxic comments, such as hate speech, threats, and insults, can help in keeping discussions fruitful.
In addition, new regulations in certain European countries have been established enforcing to delete illegal content in less than 72 hours.\footnote{\url{https://www.bbc.com/news/technology-42510868}}

Active research on the topic deals with common challenges of natural language processing, such as long-range dependencies or misspelled and idiosyncratic words. 
Proposed solutions include bidirectional recurrent neural networks with attention~\cite{Pavlopoulos:17} and the use of pretrained word embeddings~\cite{Badjatiya:17}.
However, many classifiers suffer from insufficient variance in methods and training data and therefore often tend to fail on the long tail of real world data~\cite{Zhang:18}.
For future research, it is essential to know which challenges are already addressed by state-of-the-art classifiers and for which challenges current solutions are still error-prone.

We take two datasets into account to investigate these errors: comments on Wikipedia talk pages presented by Google Jigsaw during Kaggle's Toxic Comment Classification Challenge\footnote{\url{https://www.kaggle.com/c/jigsaw-toxic-comment-classification-challenge}} and a Twitter Dataset by~\citet{Davidson:17}. 
These sets include common difficulties in datasets for the task: They are labeled based on different definitions; they include diverse language from user comments and Tweets; and they present a multi-class and a multi-label classification task respectively.

On these datasets we propose an ensemble of state-of-the-art classifiers. 
By analysing false negatives and false positives of the ensemble we get insights about open challenges that all of the approaches share.
Therefore, our main contributions are:

1) We are the first to apply and compare a range of strong classifiers to a new public multi-label dataset of more than 200,000 user comments. 
Each classifier, such as Logistic Regression, bidirectional RNN and CNN, is meant to tackle specific challenges for text classification. 
We apply the same classifiers to a dataset of Tweets to validate our results on a different domain.

2) We apply two different pretrained word embeddings for the domain of user comments and Tweets to compensate errors such as idiosyncratic and misspelled words.

3) We compare the classifiers' predictions and show that they make different errors as measured by Pearson correlation coefficients and F1-measures.
Based on this, we create an ensemble that improves macro-averaged F1-measure especially on sparse classes and data with high \mbox{variance.}

4) We perform a detailed error analysis on results of the ensemble.
The analysis points to common errors of all current approaches.
We propose directions for future work based on these unsolved challenges.


\section{Related Work}
\label{sec:related-work}

\paragraph{Task definitions.} 
Toxic comment classification is not clearly distinguishable from its related tasks.
Besides looking at toxicity of online comments~\cite{Wulczyn:16,Georgakopoulos:18}, related research includes the investigation of hate speech~\cite{Badjatiya:17,Burnap:16,Davidson:17,Gambaeck:17,Gitari:15,Schmidt:17,Vigna:17,Warner:12}, online harassment~\cite{Yin:09,Golbeck:17}, abusive language~\cite{Mehdad:16,Park:17}, cyberbullying~\cite{Dadvar:13,Dinakar:12,VanHee:15,Zhong:16} and offensive language~\cite{Chen:12,Xiang:12}.  
Each field uses different definitions for their classification, still similar methods can often be applied to different tasks. 
In our work we focus on toxic comment detection and show that the same method can effectively be applied to a hate speech detection task.

\paragraph{Multi-class approaches.} 
Besides traditional binary classification tasks, related work considers different aspects of toxic language, such as ``racism''~\cite{Greevy:04,Waseem:16,Kwok:13} and ``sexism''~\cite{WaseemHovy:16,Jha:17}, or the severity of toxicity~\cite{Davidson:17,Sharma:18}. 
These tasks are framed as multi-class problems, where each sample is labeled with exactly one class out of a set of multiple classes. 
The great majority of related research considers only multi-class problems. 
This is remarkable, considering that in real-world scenarios toxic comment classification can often be seen as a multi-label problem, with user comments fulfilling different predefined criteria at the same time. 
We therefore investigate both a multi-label dataset containing six different forms of toxic language and a multi-class dataset containing three mutually exclusive classes of toxic language.

\paragraph{Shallow classification and neural networks.} 
Toxic comment identification is a supervised classification task and approached by either methods including manual feature engineering~\cite{Burnap:15,Mehdad:16,Waseem:16,Davidson:17,Nobata:16,Kennedy:17,Samghabadi:17,Robinson:18} or the use of (deep) neural networks~\cite{Ptaszynski:17,Pavlopoulos:17,Badjatiya:17,Vigna:17,Park:17,Gambaeck:17}. 
While in the first case manually selected features are combined into input vectors and directly used for classification, neural network approaches are supposed to automatically learn abstract features above these input features. 
Neural network approaches appear to be more effective for learning~\cite{Zhang:18}, while feature-based approaches preserve some sort of explainability. We focus in this paper on baselines using deep neural networks (e.g.\ CNN and Bi-LSTM) and shallow learners, such as Logistic Regression approaches on word n-grams and character n-grams.

\paragraph{Ensemble learning.} 
\citet{Burnap:15} studied advantages of ensembles of different classifiers. 
They combined results from three feature-based classifiers. 
Further the combination of results from Logistic Regression and a Neural Network has been studied~\cite{Gao:17,risch2018aggression}.
\citet{Zimmerman:18} investigated ensembling models with different hyper-parameters.
To our knowledge, the approach presented in this paper, combining both various model architectures and different word embeddings for toxic comment classification, has not been investigated so far.

\section{Datasets and Tasks}
\label{sec:datasets}

The task of toxic comment classification lacks a consistently labeled standard dataset for comparative evaluation~\cite{Schmidt:17}. 
While there are a number of annotated public datasets in adjacent fields, such as hate speech~\cite{Ross:16,Gao:17}, racism/sexism~\cite{Waseem:16,WaseemHovy:16} or harassment~\cite{Golbeck:17} detection, most of them follow different definitions for labeling and therefore often constitute different problems.
\begin{table}[t!]
\begin{center}
\begin{tabular}{lr} 
\toprule
Class & \# of occurrences \\ 
\midrule 
Clean & 201,081\\
Toxic & 21,384\\
Obscene & 12,140\\
Insult & 11,304\\
Identity Hate & 2,117\\
Severe Toxic & 1,962\\
Threat & 689\\
\bottomrule
\end{tabular}
\end{center}
\caption{\label{tab:kaggle} Class distribution of Wikipedia dataset. The distribution shows a strong class imbalance. }
\end{table}
\begin{table}[t!]
\begin{center}
\begin{tabular}{lr} 
\toprule
Class & \# of occurrences \\ 
\midrule 
Offensive & 19,190\\
Clean & 4,163\\
Hate & 1,430\\
\bottomrule
\end{tabular}
\end{center}
\caption{\label{tab:davidson} Class distribution of Twitter dataset. The majority class of the dataset consists of offensive Tweets. }
\end{table}

\subsection{Wikipedia Talk Pages dataset}

We analyse a dataset published by Google Jigsaw in December 2017 over the course of the `Toxic Comment Classification Challenge' on Kaggle.
It includes 223,549 annotated user comments collected from Wikipedia talk pages and is the largest publicly available for the task.
These comments were annotated by human raters with the six labels `toxic', `severe toxic, `insult', `threat', `obscene' and `identity hate'.
Comments can be associated with multiple classes at once, which frames the task as a multi-label classification problem.
Jigsaw has not published official definitions for the six classes. 
But they do state that they defined a toxic comment as ``a rude, disrespectful, or unreasonable comment that is likely to make you leave a discussion''.\footnote{\url{http://www.perspectiveapi.com/}}

The dataset features an unbalanced class distribution, shown in Table~\ref{tab:kaggle}. 
201,081 samples fall under the majority `clear' class matching none of the six categories, whereas 22,468 samples belong to at least one of the other classes. 
While the `toxic' class includes 9.6\% of the samples, only 0.3\% are labeled as `threat', marking the smallest class.

Comments were collected from the English Wikipedia and are mostly written in English with some outliers, e.g., in Arabic, Chinese or German language. 
The domain covered is not strictly locatable, due to various article topics being discussed. 
Still it is possible to apply a simple categorization of comments as follows:\footnote{Disclaimer: This paper contains examples that may be considered profane, vulgar, or offensive. These contents do not reflect the views of the authors and exclusively serve to explain linguistic research challenges.}
1) `community-related':\\[.3ex]
\noindent\begin{tabular}{|p{.95\columnwidth}|}
\hline
Example: \textit{``If you continue to vandalize\Tstrut\newline Wikipedia, you will be blocked from editing.''}\\
\hline
\end{tabular}\\[1ex]
2) `article-related':\\[.3ex]
\noindent\begin{tabular}{|p{.95\columnwidth}|}
\hline
Example: \textit{``Dark Jedi Miraluka from the Mid-\Tstrut\newline Rim world of Katarr, Visas Marr is the lone surviving member of her species.''}\\
\hline
\end{tabular}\\[1ex]
3) `off-topic':\\[.3ex]
\noindent\begin{tabular}{|p{.95\columnwidth}|}
\hline
Example: \textit{``== I hate how my life goes today\Tstrut\newline == Just kill me now….''}\\
\hline
\end{tabular}\\[1ex]
\subsection{Twitter dataset}

Additionally we investigate a dataset introduced by~\citet{Davidson:17}. 
It contains 24,783 Tweets fetched using the Twitter API and annotated by CrowdFlower workers with the labels `hate speech', `offensive but not hate speech' and  `neither offensive nor hate speech'. 
Table~\ref{tab:davidson} shows the class distribution. 
We observe a strong bias towards the ‘offensive’ class making up 77.4\% of the comments caused by sampling  tweets by seed keywords from Hatebase.org.
We choose this dataset to show that our method is also applicable to multi-class problems and works with Tweets, which usually have a different structure than other online user comments due to character limitation.

\subsection{Common Challenges}

We observe three common challenges for Natural Language Processing in both datasets:

\begin{description}
\item[Out-of-vocabulary words.]
A common problem for the task is the occurrence of words that are not present in the training data. These words include slang or misspellings, but also intentionally obfuscated content.
\item[Long-Range Dependencies.] 
The toxicity of a comment often depends on expressions made in early parts of the comment. 
This is especially problematic for longer comments (\textgreater50 words) where the influence of earlier parts on the result can vanish. 
\item[Multi-word phrases.] 
We see many occurrences of multi-word phrases in both datasets. 
Our algorithms can detect their toxicity only if they can recognize multiple words as a single (typical) hateful phrase. 
\end{description}

\section{Methods and Ensemble}
\label{sec:methods}

In this section we study baseline methods for the above mentioned common challenges.
Further, we propose our ensemble learning architecture.
Its goal is to minimize errors by detecting optimal methods for a given comment.

\subsection{Logistic Regression}

The Logistic Regression (LR) algorithm is widely used for binary classification tasks. 
Unlike deep learning models, it requires manual feature engineering. 
Contrary to Deep Learning methods, LR permits obtaining insights about the model, such as observed coefficients. 
Research from~\citet{WaseemHovy:16} shows that word and character n-grams belong to one of the most indicative features for the task of hate speech detection. 
For this reason we investigate the use of word and character n-grams for LR models.

\subsection{Recurrent Neural Networks}

\begin{table*}[t!]
\begin{center}
\begin{tabular}{lrrrrrrrr} 
\toprule
Model & \multicolumn{4}{c}{Wikipedia} & \multicolumn{4}{c}{Twitter} \\ 
\midrule 
 & P & R & F1 & AUC & P & R & F1 & AUC \\
 CNN (FastText) & .73 & .86 & .776 & .981 & .73 & .83 & .775 & .948 \\
 CNN (Glove) & .70 & .85 & .748 & .979 & .72 & .82 & .769 & .945 \\
 LSTM (FastText)& .71 & .85 & .752 & .978 & .73 & .83 & .778 & .955 \\
 LSTM (Glove)& \textbf{.74} & .84 & .777 & .980 & .74 & .82 & .781 & .953 \\
 Bidirectional LSTM (FastText)& .71 & .86 & .755 & .979 & .72 & .84 & .775 & .954 \\
 Bidirectional LSTM (Glove)& \textbf{.74} & .84 & .777 & .981 & .73 & \textbf{.85} & .783 & .953 \\
 Bidirectional GRU (FastText)& .72 & .86 & .765 & .981 & .72 & .83 & .773 & .955 \\
 Bidirectional GRU (Glove)& .73 & .85 & .772 & .981 & .76 & .81 & .784 & .955 \\
 Bidirectional GRU Attention (FastText)& \textbf{.74} & \textbf{.87} & \textbf{.783} & \textbf{.983} & .74 & .83 & \textbf{.791} & \textbf{.958} \\
 Bidirectional GRU Attention (Glove)& .73 & \textbf{.87} & .779 & \textbf{.983} & \textbf{.77} & .82 & \textbf{.790} & .952 \\
 Logistic Regression (char-ngrams)& \textbf{.74} & .84 & .776 & .975 & .73 & .81 & .764 & .937 \\
 Logistic Regression (word-ngrams)& .70 & .83 & .747 & .962 & .71 & .80 & .746 & .933 \\
\midrule
 Ensemble & \textbf{.74} & \textbf{.88} &  \textbf{.791} & \textbf{.983} & .76 & .83 &  \textbf{.793} & .953 \\
\bottomrule
\end{tabular}
\end{center}
\caption{\label{tab:results} Comparison of precision, recall, F1-measure, and ROC AUC on two datasets. The results show that the ensemble outperforms the individual classifiers in F1-measure. The strongest individual classifier on both datasets is a bidirectional GRU network with attention layer.}
\end{table*}

Recurrent Neural Networks (RNNs) interpret a document as a sequence of words or character n-grams.
We use four different RNN approaches: An LSTM (Long-Short-Term-Memory Network), a bidirectional LSTM, a bidirectional GRU (Gated Recurrent Unit) architecture and a bidirectional GRU with an additional attention layer.

\paragraph{LSTM.}
Our LSTM model takes a sequence of words as input. 
An embedding layer transforms one-hot-encoded words to dense vector representations and a spatial dropout, which randomly masks 10\% of the input words, makes the network more robust.
To process the sequence of word embeddings, we use an LSTM layer with 128 units, followed by a dropout of 10\%.
Finally, a dense layer with a sigmoid activation makes the prediction for the multi-label classification and a dense layer with softmax activation makes the prediction for the multi-class classification.

\paragraph{Bidirectional LSTM and GRU.}
Bidirectional RNNs can compensate certain errors on long range dependencies. 
In contrast to the standard LSTM model, the bidirectional LSTM model uses two LSTM layers that process the input sequence in opposite directions.
Thereby, the input sequence is processed with correct and reverse order of words. 
The outputs of these two layers are averaged. 
Similarly, we use a bidirectional GRU model, which consists of two stacked GRU layers. 
We use layers with 64 units.
All other parts of the neural network are inherited from our standard LSTM model. 
As a result, this network can recognize signals on longer sentences where neurons representing words further apart from each other in the LSTM sequence will `fire' more likely together.

\paragraph{Bidirectional GRU with Attention Layer.}
\citet{Gao:17} phrase that ``attention mechanisms are suitable for identifying specific small regions indicating hatefulness in long comments''.
In order to detect these small regions in our comments, we add an attention layer to our bidirectional GRU-based network following the work of~\citet{Yang:16}.

\subsection{Convolutional Neural Networks}

Convolutional Neural Networks (CNNs) are recently becoming more popular for text classification tasks. 
By intuition they can detect specific combinations of features, while RNNs can extract orderly information~\cite{Zhang:18}. 
On character level, CNNs can deal with obfuscation of words. 
For our model we choose an architecture comparable to the approach of~\citet{Kim:14}.

\subsection{(Sub-)Word Embeddings}

Using word embeddings trained on very large corpora can be helpful in order to capture information that is missing from the training data~\cite{Zhang:18}.
Therefore we apply Glove word embeddings trained on a large Twitter corpus by \citet{Pennington:14}.
In addition, we use sub-word embeddings as introduced by~\citet{Bojanowski:17} within the FastText tool.
The approach considers substrings of a word to infer its embedding.
This is important for learning representations for misspelled, obfuscated or abbreviated words which are often present in online comments.
We train FastText embeddings on 95 million comments on Wikipedia user talk pages and article talk pages.\footnote{\url{https://figshare.com/articles/Wikipedia_Talk_Corpus/4264973}}
We apply the skip-gram method with a context window size of 5 and train for 5 epochs.

\subsection{Ensemble Learning}

Each classification method varies in its predictive power and may conduct specific errors. 
For example, GRUs or LSTMs may miss long range dependencies for very long sentences with 50 or more words but are powerful in capturing phrases and complex context information.
Bi-LSTMs and attention based networks can compensate these errors to a certain extent.
Subword Embeddings can model even misspelled or obfuscated words.

Therefore, we propose an ensemble deciding which of the single classifiers is most powerful on a specific kind of comment.
The ensemble observes features in comments, weights and learns an optimal classifier selection for a given feature combination.
For achieving this functionality, we observe the set of out-of-fold predictions from the various approaches and train an ensemble with gradient boosting decision trees.
We perform 5-fold cross-validation and average final predictions on the test set across the five trained models.

\section{Experimental Study}
\label{sec:experiments}


Our hypothesis is that the ensemble learns to choose an optimal combination of classifiers based on a set of comment features. 
Because the classifiers have different strengths and weaknesses, we expect the ensemble to outperform each individual classifier.
Based on results from previous experiments mentioned in Section~\ref{sec:related-work} we expect that the state-of-the-art models have a comparable performance and none outperforms the others significantly. 
This is important because otherwise the ensemble learner constantly prioritizes the outperforming classifier.
We expect our ensemble to perform well on both online comments and Tweets despite their differing language characteristics such as comment length and use of slang words.

\subsection{Setup}

To evaluate our hypotheses, we use the following setup: We compare six methods from Section~\ref{sec:methods}. For the neural network approaches we apply two different word embeddings each and for LR we use character and word n-grams as features.

We need binary predictions to calculate precision, recall and the resulting F1-measure. 
To translate the continuous sigmoid output for the multi-label task (Wikipedia dataset) into binary labels we estimate appropriate threshold values per class. 
For this purpose we perform a parameter search for the threshold to optimize the F1-measure using the whole training set as validation. 
In case of the multi-class task (Twitter dataset) the softmax layer makes the parameter search needless, because we can simply take the label with the highest value as the predicted one.

We choose the macro-average F1 measure since it is more indicative than the micro-average F1 for strongly unbalanced datasets~\citep{Zhang:18}.
For the multi-label classification we measure macro-precision and -recall for each class separately and average their results to get the F1-measure per classifier.
The Area under the Receiver Operating Curve (ROC AUC) gives us a measurement of classifier performance without the need for a specific threshold.
We add it to provide additional comparability of the results.

\subsection{Correlation Analysis}

\begin{table}[t!]
\begin{center}
\begin{tabular}{lrrr} 
\toprule
Class & \multicolumn{2}{c}{F1} & Pearson \\ 
\midrule 
& \multicolumn{3}{c}{\textbf{Different word embeddings}} \\
& GRU+G & GRU+FT & \\
W avg. & .78 & .78& .95 \\
W threat & .70 & .69& .92\\
T avg. & .79& .79& .96\\
T hate & .53& .54& .94\\
& CNN+G & CNN+FT & \\
W avg. & .75& .78& .91\\
\textbf{W threat} & \textbf{.67} & \textbf{.73}& \textbf{.82}\\
T avg. & .77&.78 & .94\\
T hate & .49& .53& .90\\
\midrule 
& \multicolumn{3}{c}{\textbf{Different NN architectures}} \\
& CNN & BiGRU Att & \\
\textbf{W avg.} & \textbf{.78}& \textbf{.78}& \textbf{.85}\\
\textbf{W threat} & \textbf{.73}& \textbf{.71}& \textbf{.65}\\
T avg. & .78& .79& .96\\
T hate & .50& .49& .93\\
\midrule
& \multicolumn{3}{c}{\textbf{Shallow learner and NN}} \\
& CNN & LR char & \\
\textbf{W avg.} & \textbf{.78}& \textbf{.78}& \textbf{.86}\\
\textbf{W threat} & \textbf{.73}& \textbf{.74}& \textbf{.78}\\
T avg. & .78& .76& .92\\
\textbf{T hate} & \textbf{.50}& \textbf{.51}& \textbf{.86}\\
& BiGRU Att & LR char & \\
\textbf{W avg.} & \textbf{.78}& \textbf{.78}& \textbf{.84}\\
\textbf{W threat} & \textbf{.71}& \textbf{.74}& \textbf{.67}\\
T avg. & .79& .76& .92\\
T hate & .49& .51& .88\\
\midrule 
& \multicolumn{3}{c}{\textbf{Character- and word-ngrams}} \\
& LR word & LR char & \\
\textbf{W avg.} & \textbf{.75}& \textbf{.78}& \textbf{.83}\\
\textbf{W threat} & \textbf{.70}& \textbf{.74}& \textbf{.69}\\
T avg. & .75& .77& .94\\
T hate & .50& .51& .91\\
\bottomrule
\end{tabular}
\end{center}
\caption{F1-measures and Pearson correlations of different combinations of classifiers. When the pearson score is low and F1 is similar, an ensemble performs best.
We see that this appears mostly on the Wikipedia dataset and on the respective minority classes `threat' and `hate'. 
`W': Wikipedia dataset; `T': Twitter dataset;  `G': Glove embeddings; `FT': FastText embeddings; `avg.': Averaged}
\label{tab:correlation}
\vspace{-15pt}
\end{table}

Total accuracy of the ensemble can only improve when models with comparable accuracy produce uncorrelated predictions.
We therefore measure the correlation of the predictions of different classifiers.
We look at a set of combinations, such as shallow learner combined with a neural net, and inspect their potential for improving the overall prediction.
For measuring the disparity of two models we use the Pearson correlation coefficient. 
The results are shown in Table~\ref{tab:correlation}.

\subsection{Experimental Results}

As shown in Table~\ref{tab:results} our ensemble outperforms the strongest individual method on the Wikipedia dataset by approximately one percent F1-measure. 
We see that the difference in F1-measure between the best individual classifier and the ensemble is higher on the Wikipedia dataset as on the Twitter dataset.
This finding is accompanied by the results in Table~\ref{tab:correlation} which show that most classifier combinations present a high correlation on the Twitter dataset and are therefore less effective on the ensemble.
An explanation for this effect is that the text sequences within the Twitter set show less variance than the ones in the Wikipedia dataset. 
This can be reasoned from 1) their sampling strategy based on a list of terms, 2) the smaller size of the dataset and 3) less disparity within the three defined classes than in the six from the Wikipedia dataset.
With less variant data one selected classifier for a type of text can be sufficient.

As the results in Table~\ref{tab:correlation} show, ensembling is especially effective on the sparse classes ``threat'' (Wikipedia) and ``hate'' (Twitter).
The predictions for these two classes have the weakest correlation.
This can be exploited when dealing with strongly imbalanced datasets, as often the case in toxic comment classification and related tasks.
Table~\ref{tab:correlation} gives us indicators for useful combinations of classifiers. 
Combining our shallow learner approach with Neural Networks is highly \mbox{effective.}
Contrary to that we see that the different word embeddings used do not lead to strongly \mbox{differing predictions.} 
Another finding is that word and character n-grams learned by our Logistic Regression classifier produce strongly uncorrelated predictions that can be combined for increasing accuracy.

\section{Detailed Error Analysis}
\label{sec:error-analysis}

The ensemble of state-of-the-art classifiers still fails to reach F1-measures higher than 0.8. 
To find out the remaining problems we perform an extensive error analysis on the result of the ensemble.

We analyse common error classes of our ensemble based on research from~\citet{Zhang:18,Zhang2:18,Qian:18,Davidson:17,Schmidt:17,Nobata:16}. 
Moreover, we add additional error classes we encountered  during our manual analysis. 
To address deficits in both precision and recall we inspect false negative and false positive classifications.  
We focus on error classes with the highest frequency in the observed samples.
The occurrence of an error class within a comment is taken to be binary (occurs in comment or not).

We present the results on class `toxic' of the Wikipedia dataset and class `hate' of the Twitter dataset.
Both classes are of high significance for the task of user comment moderation.
Our ensemble results in 1794 false negatives and 1581 false positives for the Wikipedia dataset.
We choose 200 random samples out of each set as representatives. 
For the smaller Twitter dataset we get 55 false negatives and 58 false positives, we perform our analysis on all of these samples.

\subsection{Error Classes of False Negatives}
\label{sec:false-negatives}

\paragraph{Doubtful labels.} 
We observe a high number of comments for which we question the original label when taking the respective class definition into account. 
A common occurrence is actual toxic or hateful content that is cited by the comment's author.
Another pattern is the use of potentially toxic words within an explanation or self reproach.\\[.3ex]
\noindent\begin{tabular}{|p{.95\columnwidth}|}
\hline
Example: \textit{``No matter how upset you may be\Tstrut\newline there is never a reason to refer to another editor as `an idiot'\ ''}\\
\hline
\end{tabular}\\[1ex]
We find that 23\% of sampled comments in the false negatives of the Wikipedia dataset do not fulfill the toxic definition in our view. 
Taking the hate speech definition of the authors into account, we question 9\% of the Twitter dataset samples. For the remaining error classes we only include the comments with undoubtful labels.

\paragraph{Toxicity without swear words.} 
\citet{Davidson:17} phrase the problem that hate speech may not contain hate or swear words at all.\\[.3ex]
\noindent\begin{tabular}{|p{.95\columnwidth}|}
\hline
Example: \textit{``she looks like a horse''}\Tstrut\\
\hline
\end{tabular}\\[1ex]
50\% of Wikipedia dataset samples have no common hate or swear word in them. 
This makes it the largest error class for the Wikipedia dataset and shows that our classifiers often fail when there are no obvious hateful words present.
We observe this in 18\% of hate speech comments from the Twitter dataset. 
It is important to notice that the frequency of swear words is naturally higher within this dataset, because of its sampling method with hateful words as seeds. 
In many cases the problem is a lack of paradigmatic context. 
Hence, an important research topic for future work is investigating improved semantic embeddings, which can better distinguish different paradigmatic contexts.

\paragraph{Rhetorical questions.} 
It is common practice to wrap toxic statements online within rhetorical or suggestive questions as pointed out by~\citet{Schmidt:17}.\\[.3ex]
\noindent\begin{tabular}{|p{.95\columnwidth}|}
\hline
Example: \textit{``have you no brain?!?!''}\Tstrut\\
\hline
\end{tabular}\\[1ex]
21\% of Wikipedia dataset samples and 10\% of Twitter dataset samples contain a rhetorical or suggestive question. Again paradigmatic context can help to identify this kind of comments. 
An additional signal is the existence of question words and question marks.

\paragraph{Metaphors and comparisons.} 
Subtle metaphors and comparisons often require understanding of implications of language or additional world knowledge. 
\citet{Zhang:18} and \citet{Schmidt:17} report on this common error class.\\[.3ex] 
\noindent\begin{tabular}{|p{.95\columnwidth}|}
\hline
Example: \textit{``Who are you a sockpuppet for?''}\Tstrut\\
\hline
\end{tabular}\\[1ex]
We only see this problem in the Wikipedia dataset samples with 16\% of false negatives impacted.

\paragraph{Idiosyncratic and rare words.}
Errors caused by rare or unknown words are reported by~\citet{Nobata:16,Zhang:18,Qian:18}. 
From our observation they include misspellings, neologisms, obfuscations, abbreviations and slang words. 
Even though some of these words appear in the embedding, their frequency may be too low to correctly detect their meaning on our word embeddings.\\[.3ex]
\noindent\begin{tabular}{|p{.95\columnwidth}|}
\hline 
Example: \textit{``fucc nicca yu pose to be pullin up''}\Tstrut\\
\hline
\end{tabular}\\[1ex]
We find rare or unknown words in 30\% of examined false negatives from the Wikipedia dataset and in 43\% of Twitter dataset samples.
This also reflects the common language on Twitter with many slang words, abbreviations and misspellings. One option to circumvent this problem is to train word embeddings on larger corpora with even more variant language.

\paragraph{Sarcasm and irony.} 
\citet{Nobata:16} and \citet{Qian:18} report the problem of sarcasm for hate speech detection. 
As sarcasm and irony detection is a hard task itself, it also increases difficulty of toxic comment classification, because the texts usually state the opposite of what is really meant.\\[.2ex]
\noindent\begin{tabular}{|p{.95\columnwidth}|}
\hline
Example: \textit{``hope you're proud of yourself.\Tstrut\newline Another milestone in idiocy.''}\\
\hline
\end{tabular}\\[1ex]
Sarcasm or irony appears in 11\% of Wikipedia dataset samples, but in none of the Twitter dataset samples.

\subsection{Error Classes of False Positives}
\paragraph{Doubtful labels.} 
We find that 53\% of false positive samples from the Wikipedia dataset actually fall under the definition of toxic in our view, even though they are labeled as non-toxic.
Most of them contain strong hateful expressions or spam. 
We identify 10\% of the Twitter dataset samples to have questionable labels.\\[.3ex]
\noindent\begin{tabular}{|p{.95\columnwidth}|}
\hline
Example: \textit{``IF YOU LOOK THIS UP UR A\Tstrut\newline DUMB RUSSIAN''}\\
\hline
\end{tabular}\\[1ex]
The analysis show that doubtful labels belong to one of the main reasons for a false classification on the Wikipedia dataset, especially for the false positives. 
The results emphasize the importance of taking labeler agreement into account when building up a dataset to train machine learning models. It also shows the need for clear definitions especially for classes with high variance like toxicity. Besides that, a deficient selection of annotators can amplify such problems as \citet{Waseem:18} point out.

\paragraph{Usage of swear words in false positives.} 
Classifiers often learn that swear words are strong indicators for toxicity in comments. 
This can be problematic when non-toxic comments contain such terms. 
\citet{Zhang:18} describe this problem as dealing with non distinctive features.\\[.3ex]
\noindent\begin{tabular}{|p{.95\columnwidth}|}
\hline
Example: \textit{``Oh, I feel like such an asshole now.\Tstrut\newline Sorry, bud.''}\\
\hline
\end{tabular}\\[1ex]
60\% of false positive Wikipedia dataset samples and 77\% of Twitter dataset samples contain swear words.
In this case, the paradigmatic context is not correctly distinguished by the embedding. 
Hence, the classifier considered signals for the trigger word (a swear word) stronger, than other signals from the context, here a first person statement addressing the author himself.

\paragraph{Quotations or references.}
We add this error class because we observe many cases of references to toxic or hateful language in actual non-hateful comments.\\[.3ex]
\noindent\begin{tabular}{|p{.95\columnwidth}|}
\hline
Example: \textit{``I deleted the Jews are dumb\Tstrut\newline comment.''}\\
\hline
\end{tabular}\\[1ex]
In the Wikipedia dataset samples this appears in 17\% and in the Twitter dataset in 8\% of comments. Again the classifier could not recognize the additional paradigmatic context referring to typical actions in a forum, here explicitly expressed with words `I deleted the\ldots ' and ` \ldots comment'.

\paragraph{Idiosyncratic and rare words.} 
Such words (as described in Section~\ref{sec:error-analysis}) in non-toxic or non-hateful comments cause problems when the classifier misinterprets their meaning or when they are slang that is often used in toxic language.\\[.3ex]
\noindent\begin{tabular}{|p{.95\columnwidth}|}
\hline
Example: \textit{``WTF man. Dan Whyte is Scottish''}\Tstrut\\
\hline
\end{tabular}\\[1ex]
8\% of Wikipedia dataset samples include rare words. 
In the Twitter dataset sample the frequency is higher with 17\%, but also influenced by common Twitter language.

\section{Conclusion}
\label{sec:conclusion}
In this work we presented multiple approaches for toxic comment classification.
We showed that the approaches make different errors and can be combined into an ensemble with improved F1-measure.
The ensemble especially outperforms when there is high variance within the data and on classes with few examples.
Some combinations such as shallow learners with deep neural networks are especially effective.
Our \mbox{error analysis} on results of the ensemble identified difficult subtasks of toxic comment classification. 
We find that a large source of errors is the lack of consistent quality of labels.
Additionally most of the unsolved challenges occur due to missing training data with highly idiosyncratic or rare vocabulary. 
Finally, we suggest further research in representing world knowledge with embeddings to improve distinction between paradigmatic contexts.

\section*{Acknowledgement}
Our work is funded by the European Union’s Horizon 2020 research and innovation programme under grant agreement No. 732328 (FashionBrain) and by the German Federal Ministry of Education and Research (BMBF) under grant agreement No. 01UG1735BX (NOHATE).

\bibliography{emnlp2018}
\bibliographystyle{acl_natbib_nourl}

\end{document}